%% file: main.tex
\documentclass[runningheads]{llncs}
\usepackage{amsmath}
\usepackage[T1]{fontenc}
\usepackage{graphicx,verbatim}
\usepackage{cuted}
\usepackage{amssymb}%
\usepackage{pifont}%
\usepackage[inkscapelatex=false]{svg}
\usepackage{booktabs}
\usepackage{subcaption}
\usepackage{hyperref}
\hypersetup{
  hidelinks
}
\usepackage{color}

\newcommand{\xpm}[1]{{\tiny$\pm#1$}}

\usepackage{xspace}
\usepackage{xcolor}

\newcommand{\para}[1]{
\subsection{#1}}

\makeatletter
\def\@onedot{\ifx\@let@token.\else.\null\fi\xspace}

\begin{document}

\title{Millimeter-wave Imaging for Anthropometric Body Measurement}

\author{Miriam Senne\inst{1,4}\orcidID{0009-0008-6995-3033} \and
Benjamin~D.~Killeen\inst{1,4}\orcidID{0000-0003-2511-7929}\and
Christoph Baur \inst{2}\and
Nassir Navab \inst{1,4}
\and
Azade Farshad \inst{1,3,4}\orcidID{0000-0002-1080-1587}
}
\authorrunning{M. Senne et al.}
\institute{Chair for Computer Aided Medical Procedures\\
Technical University of Munich, Boltzmannstr. 3, 85748 Garching, Germany
\email{\{miriam.senne,bd.killeen,nassir.navab,azade.farshad\}@tum.de}\\
\and
Rohde \& Schwarz GmbH \& Co. KG\\
\email{christoph.baur@rohde-schwarz.com}
\and
Munich Center for Machine Learning, Munich, Germany\\
\and
ELLIS Unit Helsinki, Dept. Computer Science, Aalto University, Espoo, Finland
}

\maketitle

\begin{abstract}

\input{sec/0_abstract}

\end{abstract}

\section{Introduction}
\label{sec:intro}

Anthropometric measurements, such as waist circumference, are commonly used for diagnosis risk stratification, treatment planning, and monitoring~\cite{wang2024utilizing}, for example, as biomarkers of adiposity and cardiometabolic risk. Anthropometric data are commonly acquired using manual tape measurement, three dimensional (3D) body scanning~\cite{lu2008automated,garciadurso2024automated}, photogrammetry, or monocular imaging. Tape measurements are low-cost but operator-dependent, whereas optical approaches are more precise but require minimal clothing~\cite{yan2020anthropometric,chen2025focused}. Because some individuals do not readily undress for privacy, cultural, or mobility reasons, clothing-tolerant imaging could broaden access, reduce selection bias, and enable repeated assessment in clinical settings.

\begin{figure}[t]
  \centering
  \includegraphics[width=\columnwidth]{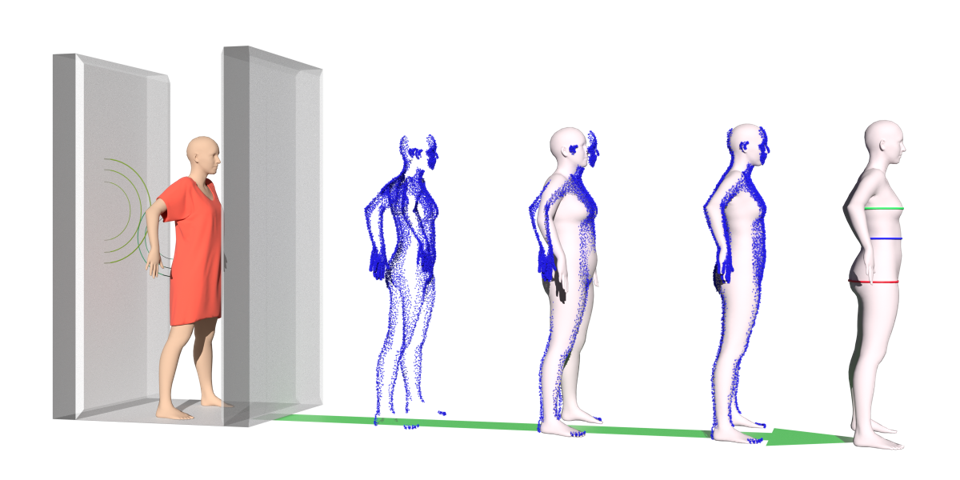}
  \caption{\textbf{Anthropometric measurement via mmWave scans.} Two planar mmWave sensor arrays are used to acquire a surface point cloud of the subject. A parametric model is fitted to the point cloud and body measurements around hip, chest and waist are computed automatically.}
  \label{fig:intro}
  \hfill
\end{figure}

Millimeter wave (mmWave) scanners offer privacy-preserving imaging that can penetrate through clothing~\cite{khalid2024emerging}. They range from sparse commercial radars that require substantial processing to recover pose structure~\cite{wu2025mmHPE,xue2021mmmesh,chen2022mmbody} to high resolution, multi-antenna holographic systems originally developed for security screening~\cite{ahmed2012advanced,sheen2002three}. Building on model fitting and registration methods from SMPLify~\cite{loper2015smpl} and 3D surface alignment based on Chamfer distance and ICP~\cite{bogo2016keep,besl1992method}, as well as recent optimization learning hybrids and probabilistic formulations~\cite{bhatnagar2020loopreg,kolotouros2021probabilistic}, this line of work motivates physics informed objectives that explicitly model sensor visibility, particularly given that RGB monocular shape estimation remains ill-posed under clothing despite strong garment modeling efforts~\cite{kanazawa2018end,kolotouros2019learning,li2022cliff}.

Here, we consider the use of mmWave scans for standardized anthropometric measurement, as shown in Fig.~\ref{fig:intro}, which to our knowledge has not been previously explored.
After processing, mmWave scans yield a partial body point cloud, with missing or unreliable regions that depend on sensor characteristics and the subject's pose, clothing, and body shape. 
We fit a parametric body model~\cite{loper2015smpl} to the point cloud, which provides a closed surface with consistent mesh topology for repeatable measurement of the chest, waist, and hip circumference and body height, following~\cite{yan2020anthropometric}.
We evaluate our approach on real mmWave scans of 27 human subjects with and without outer garments, as well as a mannequin with tape measurements and optical 3D scans as ground truth, and benchmark against an RGB monocular baseline. Overall, the results support mmWave plus parametric fitting as a clothing tolerant-approach that enables reproducible circumferential biomarkers and the potential to improve comparability across studies.

\section{Methodology}
\label{sec:method}

Given a pair of volumetric MI scans (front \& back) acquired by two planar mmWave scanners, we obtain a 3D surface of the subject's body, where it is facing the scanner panels, which we leverage for reliable extraction of anthropometric measurements, as in Fig.~\ref{fig:intro}.
Surface point clouds are extracted from each scan by identifying local reflectivity maxima along the depth direction and thresholding to obtain depth maps. After smoothing, each depth map is projected to a 3D point cloud and merged. The resulting point cloud is then downsampled to a point spacing of approximately 1 cm, which balances the need for detailed surface representation with computational efficiency.

\begin{figure}[t]
  \centering
  \includegraphics[width=\textwidth]{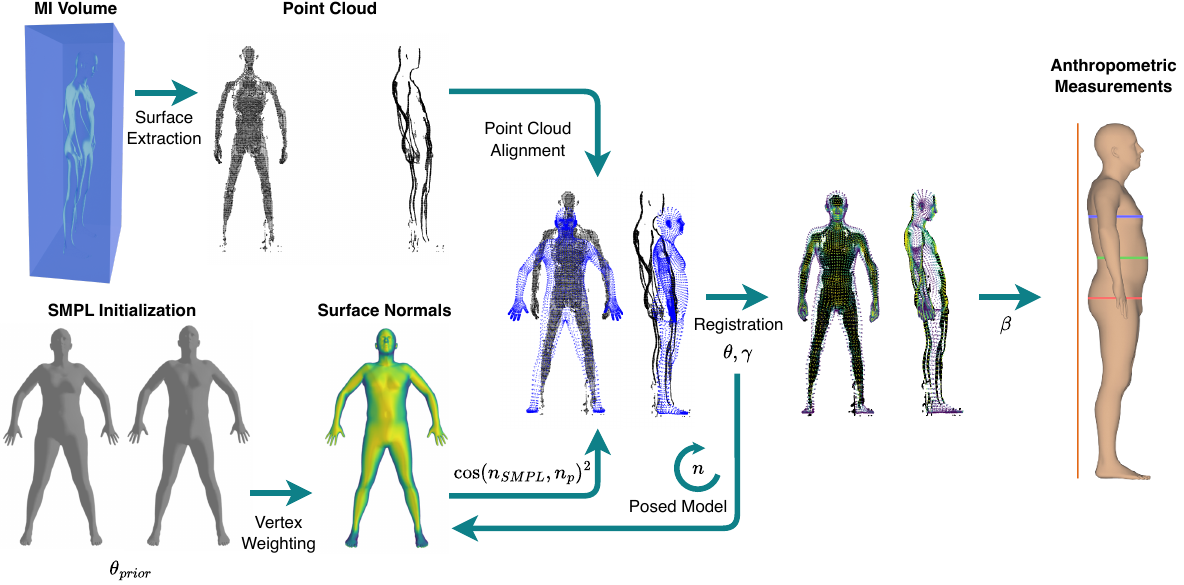}
  \caption{\textbf{Method Overview.} The MI volume is converted to a point cloud and rigidly aligned with an initialized SMPL model in A-pose. Per-vertex weights are computed from the posed SMPL normals with respect to the panel normals. The distance between SMPL vertices and point cloud is then minimized to obtain the final shape parameters $\beta$ used for anthropometric measurements.}
  \label{fig:method}
  \hfill
\end{figure}

\para{Registration}
Due to the scanner modality, regions normal to the sensor planes are densely represented while areas oriented away from both panels are not captured, necessitating shape completion to obtain a closed surface for measurement.
We use the SMPL body model as the parametric mesh representation in our pipeline. SMPL \cite{loper2015smpl} is a differentiable, skinned mesh model that maps low-dimensional shape and pose parameters to a fixed-topology triangle mesh. Shape is encoded by $\beta \in \mathrm{I\!R}^{|\beta|}$, which linearly deforms a template mesh, and pose is encoded by $\theta \in \mathrm{I\!R}^{3K}$ (K = 24 joints) as axis-angle rotations. The posed, shaped, and translated vertex positions are written $V(\beta, \theta, t) \in \mathrm{I\!R}^{N_v \times 3}$, $N_v$ being the number of vertices and $t$ being the global translation.
In our implementation, we use the standard SMPL topology ($N_v =$ 6,890) and compute per-vertex normals $n_i(\beta, \theta, t)$ from the posed, shaped and translated mesh geometry.
The registration is initialized by applying a fixed pose prior $\theta_{prior}$ to the SMPL model to provide a stable starting point for optimization. We then transform the posed mesh $V(\theta_{prior},\beta, t)$ into the point-cloud coordinate frame.
To simulate visibility to the mmWave scanner, we use the panel normal $n_{panel}$ to compute a per-vertex support weight from the squared cosine similarity between each vertex normal, $w_i(\theta, \beta) = (n_i(\theta, \beta) \cdot n_{panel})^2$.
We compute per-vertex normals $n_i$ by area-weighted averaging of adjacent face normals.
Let $F(i)$ denote the set of faces incident on vertex $i$, and $n_f$ be the normal of face $f$, then the unnormalized vertex normal is computed as
$\tilde{n}_i = \sum {f \in F(i)} n_f$,
and renormalize to obtain the unit vertex normal
$n_i = \tilde{n}_i / (\parallel\tilde{n}_i\parallel)$.
\begin{equation}
  \begin{aligned}
    E_{cham}(\theta, \beta, t)  & = \frac{1}{N_v}\sum_{i=1}w_i \space \min_{p \in P}\parallel v_i(\theta, \beta, t)-p\parallel _2^2\\
    & +\frac{1}{N_p}\sum_{j=1}\space\min_{v \in V(\theta, \beta, t)}\parallel p_j-v\parallel _2^2
  \end{aligned}
\end{equation}

Similar to other 3D full-body scanners, the surface information of the feet is low. Therefore, we also introduce an energy term which penalizes the distance of the foot soles to the ground height,
\begin{equation}
  E_{ground}(\theta, \beta, t) = \frac{1}{|F|}\sum_{i\in F}(v_i^y(\theta, \beta, t)-y_{ground})^2.
  \end{equation}

Finally, the total energy function is,
\begin{equation}
E_{total}(\theta, \beta, t) = \lambda_CE_{cham}(\theta, \beta, t)  + \lambda_GE_{ground}(\theta, \beta, t).
\end{equation}

\para{Anthropometric Body Measurement Extraction}
\label{sec:body_measurement}
We follow the ISO 8559 ~\cite{ISO8559} standard for all body measurements. The chest circumference is the horizontal measurement at the fullest part of the chest, the natural waistline is the narrowest region of the torso between the inferior margins of the ribs and the superior aspect of the hip bones, and the hip girth is measured horizontally at the level of the greatest lateral trochanteric projections while standing upright. Based on these definitions, we perform the body measurement task on the SMPL-derived human model, with fixed shape parameters and the mesh posed in an A-pose.
For each model type (male, female, neutral), we define a range of heights, expressed as fixed vertex indices, which delineate the regions over which the chest, waist, and hip circumferences are to be measured.
We compute circumference measurements by intersecting the mesh surface $S$ with a set of horizontal planes at heights $h_i$, incremented in steps of  $\Delta h = 2.5\,\mathrm{mm}$ within the region defined by lower and upper bounds $h_{min}$ and $h_{max}$. For each axial plane at height $h_i$, we extract the set of intersection points $I_i$ between the torso mesh and the plane $I_i = S \cap \{(x,y,z) : z = h_i\}$.
To simulate a measuring tape, we compute the convex hull $H_i = \text{ConvHull}(\textbf{I}_i)$ of these points and use its perimeter as the circumference estimate, $C_i = \text{Perimeter}(H_i)$. This process is repeated for all planes within the measurement region. For the chest and hip, we report the maximum circumference measured across all slices, and for the waist, we report the minimum, aligning with ISO 8559 recommendations.

\para{Optimization}
During optimization, the per-vertex weighting is updated every 50 iterations to promote scanner-aligning model positioning while still permitting flexibility in pose through the adaptive updating. This allows the model to achieve stable convergence in the earlier phases of optimization, while finer pose adjustments are promoted in the later stages. Registration terminates if there is no improvement in the objective for 20 consecutive iterations, or after a maximum of 700 iterations and learning rate was set to 0.001. To balance the competing objectives, we set the Chamfer loss coefficient $\lambda_C$  to 1 and the foot-ground constraint coefficient $\lambda_G$ to 0.5, thereby encouraging accurate foot placement without allowing this constraint to dominate the optimization process.

\section{Evaluation}
\label{sec:experiments}

\para{Mannequin Data}
\label{sec:Mannequin}
To confirm the validity of our approach, we conduct a comparative analysis using a physical mannequin. For reference, a high-resolution 3D mesh of the mannequin was acquired using an Artec Leo handheld 3D laser scanner \cite{ArtecLeo}.
Given that the human subjects in our study were measured while wearing tight-fitting garments, we performed additional manual measurements on the mannequin both without any clothing and with the mannequin dressed in thermal wear. This allowed us to assess the impact of tight-fitting apparel on manual circumference measurements. Similarly, mmWave scans were performed in two conditions: unclothed and with the mannequin wearing a blazer. Each scenario was repeated three times, with the mannequin repositioned between scans to introduce variability in alignment. Table~\ref{tab:mannequin_results} summarizes the measurements, showing that the mmWave-based measurements are comparable to those obtained from the 3D scan, even in the presence of outer garments, and that manual tape measurements are affected by clothing, particularly for the hip circumference.

\begin{figure}[t]
  \centering
  \includegraphics[width=\textwidth]{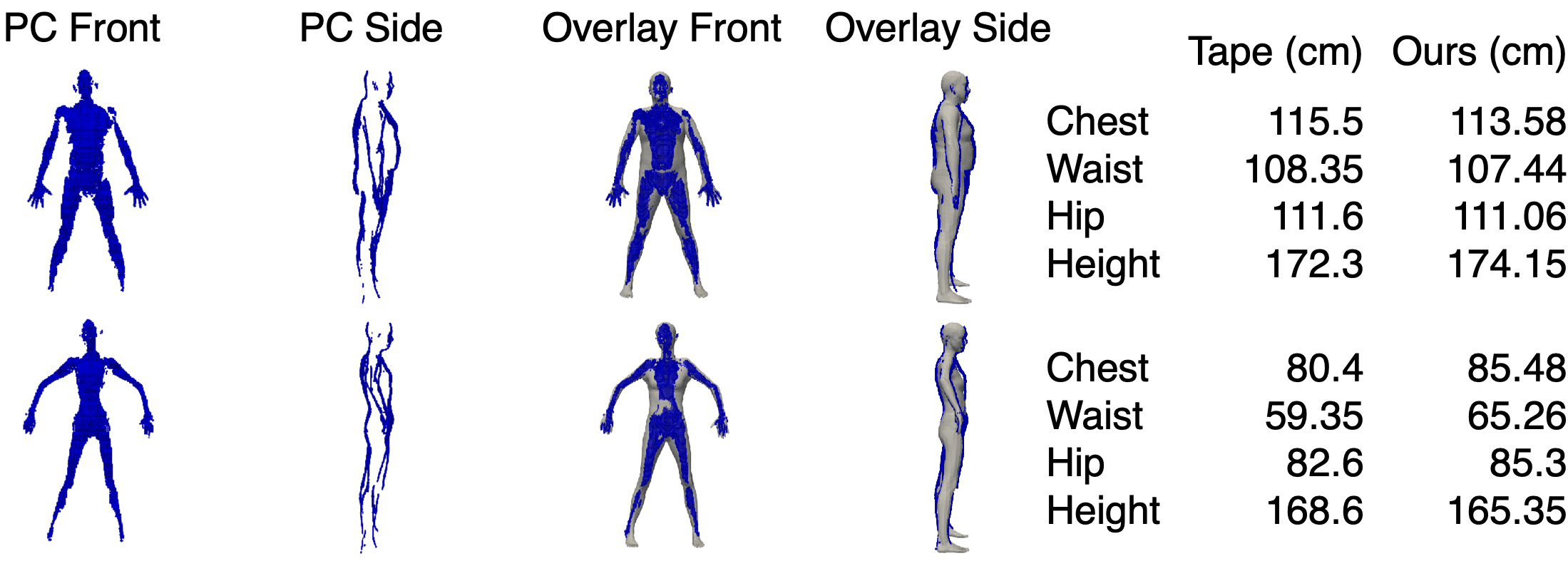}
  \caption{\textbf{Real data.} Example mmWave point clouds and SMPL fits for two participants, with mmWave- and tape-based anthropometric measurements (right).}
  \label{fig:real_data}
  \hfill
\end{figure}
\begin{table}[t]
  \caption{\textbf{Mannequin Anthropometric Measurements.} We compare the convex hull of an axial slice of the 3D scan to manual tape measurements with (C) and without (NC) clothes, to results based on SMPL fitting on mmWave data.}
  \label{tab:mannequin_results}
  \centering
\begin{tabular}{@{}lrrr@{}}
  \toprule
  & Chest Circ. (cm) & Waist Circ. (cm) & Hip Circ. (cm) \\
  \midrule
  3D Scan & 102.5 \xpm{0.8} & 81.0 \xpm{0.0} & 102.8 \xpm{0.4} \\
  $\text{Tape}_{NC}$ & 101.4 \xpm{0.4} & 79.2 \xpm{0.2} & 97.40 \xpm{0.5} \\
  $\text{Ours}_{NC}$ & 102.8 \xpm{0.4} & 80.4 \xpm{0.1} & 103.7 \xpm{0.2} \\
  $\text{Tape}_{C}$ & 102.0 \xpm{0.2} & 80.1 \xpm{0.1} & 100.4 \xpm{0.4} \\
  $\text{Ours}_{C}$ & 103.5 \xpm{1.5} & 81.0 \xpm{0.7} & 103.3 \xpm{0.3} \\
  \bottomrule
\end{tabular}
\end{table}

\para{Real Dataset}
To evaluate on real human subjects, we collected mmWave scans with paired anthropometric measurements. The cohort comprised 27 participants (18 male, 9 female). Each participant was scanned twice: once wearing regular daily clothing and once wearing tight-fitting clothing. Shoes were always removed. Anthropometric measurements were obtained by two trained personnel and included chest, waist and hip circumferences as well as body height. A horizontal laser was projected at the corresponding height of each circumference measurement to ensure consistency.
For each clothing condition, we acquired front and side-view photographs with the subject in an A-pose. These images facilitate comparison with camera-based methods and illustrate the effect of clothing on both camera-based approaches and our mmWave-based measurements.
Participant ages ranged from 21 to 45 years, and body mass index (BMI) ranged from 16.64 to 37.53 $kg/m^2$. All participants provided informed consent, and the study was approved by an ethics committee.

\begin{table}[t]
  \centering
  \caption{\textbf{Anthropometric measurement error.} Mean Absolute Deviation (MAD) in cm between tape measurements and our method based on SMPL fitting to mmWave scans, as well as a monocular RGB baseline based on STAR. $\text{SMPL}_N$ and $\text{STAR}_N$ refer to the neutral models, while $\text{SMPL}_G$ and $\text{STAR}_G$ refer to the gendered versions.}
  \label{tab:results_clothes}
  \begin{tabular}{lcccccc}
    \toprule
    Area & $\text{SMPL}_N$ & $\text{SMPL}_G$ & $\text{STAR}_N$ & $\text{STAR}_G$  & $\text{SHAPY}_{FRONT}$ & $\text{SHAPY}_{SIDE}$ \\
    \toprule
    Chest & 3.01 & 4.61 & 2.57 & \textbf{2.56} & 6.99 & 6.82 \\
    Waist & 2.02 & 1.94 & 2.24 & \textbf{1.91} & 9.65 & 9.63 \\
    Hip & \textbf{1.45} & 1.56 & 4.08 & 3.50 & 6.12 & 6.28 \\
    Height & 1.17 & 1.20 & \textbf{1.01} & 3.38 & 6.72 & 6.08 \\
    \bottomrule
  \end{tabular}
\end{table}

Table~\ref{tab:results_clothes} summarizes the Mean Absolute Deviation (MAD) between tape measurements and our method based on SMPL fitting to mmWave scans, as well as a monocular RGB baseline based on STAR. The results indicate that our mmWave-based method achieves lower MAD values across all measurement areas compared to the RGB baseline, particularly for hip and height measurements. The neutral SMPL model ($\text{SMPL}_N$) performs better than the neutral version ($\text{SMPL}_G$), while the gendered STAR model ($\text{STAR}_G$) outperforms its neutral counterpart for chest, waist and hip measurements. Overall, the mmWave-based approach demonstrates improved accuracy in anthropometric measurement compared to the RGB baseline, with some variability depending on the specific body region and model type. Fig.~\ref{fig:real_data} shows example mmWave point clouds and SMPL fits for two randomly selected participants, along with their corresponding mmWave- and tape-based anthropometric measurements.

\begin{table}[t]
  \centering
\caption{\textbf{Differences in measurements }between clothed and unclothed recordings/images. $\text{SHAPY}_\text{VIEW}$ is the same MAD over the views.}
\label{tab:clothing_results}
\begin{tabular}{lcccccl}
  \toprule
  & \multicolumn{5}{c}{Method} & \\ \toprule
  &$\text{OUR}_{\text{SMPL}_N}$&  $\text{OUR}_{\text{SMPL}_G}$&  $\text{OUR}_{\text{STAR}_G}$&    $\text{SHAPY}_\text{CLOTH}$ & $\text{SHAPY}_\text{VIEW}$ \\ \midrule
  Chest&  1.63 \xpm{1.15}&  1.7 \xpm{1.00}&  \textbf{1.30 \xpm{0.84}} &  2.88 \xpm{3.21} & 2.87 \xpm{3.51} \\
  Waist&  \textbf{0.96 \xpm{0.80}}&  1.14 \xpm{0.88} &  1.11 \xpm{0.65} & 2.47 \xpm{2.64} & 2.45 \xpm{3.52} \\
  Hip&  \textbf{1.12 \xpm{0.97}}&  1.31 \xpm{1.23}&  1.13 \xpm{0.85}
  &  1.71 \xpm{1.97} & 2.26 \xpm{2.94}\\
  Height&  \textbf{0.41 \xpm{0.24}}&  0.42 \xpm{0.39}&  0.57 \xpm{0.43} &   3.05 \xpm{2.89} & 4.69 \xpm{3.24} \\ \bottomrule
\end{tabular}
\end{table}

Table~\ref{tab:clothing_results} summarizes the differences in measurements between clothed and unclothed scans for our method and the monocular RGB baseline. The results indicate that our mmWave-based method exhibits relatively low differences between clothing conditions, particularly for waist and hip measurements, while the SHAPY method shows higher deviations, especially for chest and height measurements. This suggests that mmWave imaging is more robust to clothing compared to optical sensors, which can be significantly affected by garments, leading to less reliable anthropometric estimates under clothing conditions.

\section{Discussion}
\label{sec:conclusion}

Evaluating anthropometric measurement methods requires careful consideration of both reference standards and methodological limitations. The widespread use of manual tape measurements as ground-truth is problematic, since such methods are subject to significant inter-operator variability and positioning inaccuracies. Our results show, that tape measurements—even on a rigid mannequin—can vary by up to 0.5 cm, and even tight clothing increases circumferences by more than one centimeter. Consequently, all comparisons to tape measurements should be interpreted as general indications rather than absolute truth.

Our mmWave approach demonstrates notable robustness in reconstructing body surfaces, even in the presence of sparse or inhomogeneous point cloud coverage—challenges typical in this domain due to incomplete sensor visibility and multipath effects. The results indicate that reliable fitting and measurement can be achieved when key body regions are adequately sampled. However, misalignment arises in under-sampled regions, such as the sides or head and highlight the importance of a closed-body mesh and the dynamic vertex weighting (DVW) strategy. DVW offers consistent improvements in Chamfer Distance, particularly where standard metrics fail to penalize gaps or ``empty space,'' though the absolute differences remain modest; this nonetheless promotes greater alignment fidelity in sparse areas.

Measurement stability is also influenced by the choice of model: gendered meshes better capture sex-specific anatomy, but may be less reliable for subjects whose features do not fit typical patterns, whereas neutral models provide broader generalizability which is also discussed in the supplementary material. For chest, waist, and hip measurements, SMPL and STAR fitted to mmWave data consistently outperform SHAPY, particularly in cases with clothing or sparse coverage, where image-based methods struggle. Our mmWave pipeline—augmented by dynamic vertex weighting achieves accurate, robust body measurements that meet practical precision requirements even in challenging clothing conditions.

\section{Conclusion}

Clothing presents a fundamental limitation for optical anthropometric measurement systems, and manual tape measurements remain error-prone due to operator variability and inconsistent placement. In this work, we demonstrated that holographic mmWave imaging combined with a parametric body model fitting framework provides a viable alternative for clothing-robust anthropometry. Our results on a static mannequin show that mmWave-based measurements are comparable to those obtained from a commercial 3D scanner, even in the presence of outer garments. In real-subject experiments, the proposed framework delivers sensible measurements and exhibits greater robustness and reliability than single-view, camera-based approaches as well as manual tape measurements. These findings indicate that mmWave imaging, combined with parametric modeling, is a practical and generally applicable solution for anthropometry. Future work can potentially focus on recovering more precise subject-specific body surfaces beyond the parametric model constraint and on benchmarking the method against a broader set of imaging technologies on real subjects.

\bibliographystyle{splncs04}
\bibliography{main}

\end{document}

%% file: sec/0_abstract.tex
Body shape and circumferences are clinically informative biomarkers for risk stratification, including measures such as waist to hip ratio, limb and trunk girths, yet conventional tools such as manual tape measures and optical scanners often require undressing and sustained poses. These demands slow workflows, compromise dignity, and exclude many older adults and people with limited mobility. To make measurement fast and contactless, we leverage millimeter-wave (mmWave) radar, which preserves privacy and operates through typical clothing, enabling quick full-body acquisition.
In this work, we present a new optimization-based framework to recover 3D human shape and extract a comprehensive set of anthropometric measurements from volumetric mmWave data. Our method introduces a weighted registration pipeline that fits a parametric body model (SMPL) directly to the noisy mmWave point cloud. The core of our contribution is a vertex-weighting strategy that modulates a Chamfer energy function for reliable surface alignment and noise elimination. We further stabilize the fit by incorporating a foot-ground plane constraint and pose priors, optimizing directly for the SMPL parameters.
Together, these components enable a fast, privacy preserving workflow that delivers high fidelity body shape and measurements through clothing without cameras or disrobing and with minimal cooperation, supporting frequent risk oriented assessments in clinics and care facilities for patients of all ages and mobility levels.